\documentclass[letterpaper, 10 pt, conference]{ieeeconf}  

\IEEEoverridecommandlockouts                              

\usepackage{booktabs}
\usepackage{array}
\usepackage{multirow}
\usepackage{makecell}
\usepackage{graphicx} 
\usepackage{adjustbox}
\usepackage{graphicx} 
\usepackage{subcaption}
\usepackage{tabularx}

\usepackage{amsmath}
\usepackage{titlesec}
\usepackage{url}

\title{\LARGE \bf 
LLM+MAP: Bimanual Robot Task Planning using Large Language Models and Planning Domain Definition Language
}

\author{
{Kun Chu}$^{*}$, {Xufeng Zhao}, {Cornelius Weber}, and {Stefan Wermter}
\thanks{The authors are with the Knowledge Technology Group, Department of Informatics, University of Hamburg, 22527 Hamburg, Germany. E-mail:
        {\tt\small {\{kun.chu, xufeng.zhao, cornelius.weber, stefan.wermter\}@uni-hamburg.de}}}
\thanks{$^{*}$Corresponding author.}
}

\begin{document}
\maketitle
\thispagestyle{empty}
\pagestyle{empty}

\begin{abstract}
Bimanual robotic manipulation provides significant versatility, but also presents an inherent challenge due to the complexity involved in the spatial and temporal coordination between two hands. Existing works predominantly focus on attaining human-level manipulation skills for robotic hands, yet little attention has been paid to task planning on long-horizon timescales. With their outstanding in-context learning and zero-shot generation abilities, Large Language Models (LLMs) have been applied and grounded in diverse robotic embodiments to facilitate task planning. However, LLMs still suffer from errors in long-horizon reasoning and from hallucinations in complex robotic tasks, lacking a guarantee of logical correctness when generating the plan. Previous works, such as LLM+P, extended LLMs with symbolic planners. However, none have been successfully applied to bimanual robots.
New challenges inevitably arise in bimanual manipulation, necessitating not only effective task decomposition but also efficient task allocation.
To address these challenges, this paper introduces LLM+MAP, a bimanual planning framework that integrates LLM reasoning and multi-agent planning, automating effective and efficient bimanual task planning.
We conduct simulated experiments on various long-horizon manipulation tasks of differing complexity. Our method is built using GPT-4o as the backend, and we compare its performance against plans generated directly by LLMs, including GPT-4o, V3 and also recent strong reasoning models o1 and R1.
By analyzing metrics such as planning time, success rate, group debits, and planning-step reduction rate, we demonstrate superior performance of LLM+MAP, while also providing insights into robotic reasoning.
Code is available at \url{https://github.com/Kchu/LLM-MAP}.
\end{abstract}

\section{Introduction}

Humans effortlessly perform bimanual tasks, from tying shoelaces to flipping a book’s pages while holding a coffee cup, often without conscious thought. Our brain orchestrates complex spatial and temporal synchronizations, dynamically adjusting to external feedback in real time. In contrast, robots struggle with such dexterity. Even simple bimanual tasks require planning, synchronization, and complex computations. Bridging this gap is an inherent challenge in robotics \cite{Krebs2022, Drolet2024}, highlighting the intricate interplay of perception, control, and adaptation. 

With the development of deep learning techniques, there have been some advances in dexterous bimanual manipulations, like folding shirts \cite{shawbimanual}, cutting vegetables \cite{Grannen2023}, scooping food \cite{Grannen2022}, etc. However, most existing works concentrate on learning human-level operational skills as a whole and overlook high-level task planning. This oversight stems from the significant challenges of effectively allocating long-horizon tasks between dual robotic arms, such as task understanding and decomposition, subtask assignment, and proper execution sequencing.

Large language models (LLMs) have shown remarkable reasoning and planning abilities in diverse fields \cite{zhao2023survey} due to their training on massive textual data. Recent works have prompted LLMs in robotic task planning \cite{Ahn2022, Zhao2023, Chu2024, gao2024dag}. LLM planning requires appropriate prompting to guide LLM to generate appropriate plans for specific embodiments and ground the plans to executable actions within the task environment. This is, however, especially challenging when dealing with complex tasks. LLMs suffer from their weakness in long-horizon reasoning and hallucinations \cite{wang2024survey}, especially in spatial scenarios. Some works have explored the combination of LLMs with classical symbolic planners \cite{liu2023, xie2023translating} to alleviate the challenge. With LLMs serving as translators to formalize the natural language problems into some declarative language like PDDL (Planning Domain Definition Language, \cite{aeronautiques1998pddl}) and ASP (Answer Set Programming, \cite{lifschitz2002answer}), a plan is generated with logical correctness guarantee \cite{ding2023task, shirai2024vision}.
Building on prior successes in integrating LLMs with symbolic planning and considering the emerging challenges in bimanual robotic scenarios, we pose the following research question:
\textit{
How can LLMs be integrated with multi-agent planning to achieve efficient spatial and temporal coordination in long-horizon bimanual robotic tasks?
}

To address the question, we propose the LLM+MAP (LLM + Multi-Agent Planning with PDDL) framework, which utilizes LLMs to transform the bimanual robotic domain and problem into PDDL representation, and generates a partial-order plan through classical symbolic planners, allowing for efficient spatial and temporal coordination in bimanual manipulations. Specifically, we first define the bimanual robotic domain in PDDL, following the spatio-temporal control patterns introduced in the LABOR agent \cite{Chu2024} for the control of NICOL, a semi-humanoid robot \cite{Matthias2023}. Based on the task description in natural language with the spatial scene information from the vision-models, the LLM is then prompted to transform task configuration into PDDL representation, compatible with the domain definition we provided. Based on the planning by symbolic solvers, a partial-order plan is generated with logical correctness and executed on the bimanual robot. We conduct extensive experiments to evaluate LLM+MAP against both large-scale coding models, GPT-4o and DeepSeek-V3 \cite{liu2024deepseek}, and the state-of-the-art reasoning models, OpenAI-o1\footnote{https://openai.com/o1/} and DeepSeek-R1\footnote{In the following, we refer to DeepSeek-R1 as R1, DeepSeek-V3 as V3, and OpenAI-o1 as o1 for brevity.} \cite{deepseekr1}, on three task domains in the NICOL bimanual robot environment. Experimental results show that, compared to directly generating task plans with even strong reasoning models, our approach significantly outperforms in terms of reduced plan generation time, higher success rates, and more efficient task allocation.

\section{Related Works}
In this section, we first describe task planning methods in robotics, including symbolic planners, LLM planners, and their combined applications. Then, we present a brief overview of the works in bimanual manipulation and multi-agent planning.
\subsection{Task Planning in Robotics}
Task planning aims at generating plans, i.e., sequences of actions from a given action set, to achieve specific goals in given scenarios \cite{gerevini2020introduction, jiang2019task}. 

\noindent\textbf{Planning with Symbolic Planners.} To yield a general-purpose automated planning system, classical planners rely on a certain type of declarative language to formalize the domains and problems. Since the introduction of STRIPS \cite{fikes1971strips} in early AI research, several types of language have been proposed and widely used, including answer set programming (ASP, \cite{lifschitz2002answer}) and planning domain definition language (PDDL, \cite{aeronautiques1998pddl}). Previous works have applied symbolic planning methods in diverse tasks \cite{jiang2019task, helmert2006fast, torreno2014fmap, Ding2020, jiang2019multi}. Task and motion planning approaches employ a high-level task planner to generate symbolic actions in discrete spaces and a low-level motion planner to generate motion trajectories in continuous space \cite{kaelbling2013integrated, jiao2021efficient}. Most of the planners developed in these works would generate plans with guaranteed logical correctness, yet requiring domain-specific programming languages as domain, problem, and solution representations. Inspired by these features and recent works in \cite{liu2023, xie2023translating}, we apply PDDL to a bimanual robot scenario interacting with a human, using LLM to transform the scenario information and natural language task descriptions into PDDL representations to construct a partial-order plan using classical planners.

\noindent\textbf{Planning with LLMs.} With the rapid developments in recent years, using natural language instructions as a prompt for LLMs to directly generate task plans has become an emergent trend in robotics. Several recent methods have leveraged LLMs as planners in diverse robotic scenarios \cite{Ahn2022, huang2022language, ding2023task, Chu2024}. For instance, SayCan \cite{Ahn2022} explored utilizing LLMs to propose feasible solutions to complex tasks based on their common-sense knowledge about the world, and then ground them to specific embodiments and environments through value functions. However, LLMs suffer from their shortcomings in long-horizon reasoning and frequent hallucinations in complex tasks \cite{Zhao24EnhancingZeroshot, stechly2024self, wang2024survey}. Despite some work on iterative querying LLMs by providing feedback or error messages \cite{Wang2023, Zhao2023, Rana2024, Chu2024}, such a strategy requires effort in careful prompt and system design to handle the feedback with LLMs.

With LLMs' extraordinary in-context learning capabilities, some works investigate using LLMs to translate natural language descriptions about a task and the setting to a PDDL-readable representation, enabling classical planners to generate guaranteed solutions \cite{liu2023, xie2023translating, shirai2024vision, chen2024autotamp}. Existing approaches that integrate LLMs with symbolic planning, such as ViLaIn \cite{shirai2024vision}, focus exclusively on scenarios requiring sequential planning, without considering the requirements of parallel execution or multi-agent planning. In contrast, our work systematically examines the distinctions between sequential and parallel planning, introducing dedicated metrics to rigorously evaluate these differences.

\subsection{Bimanual Manipulation and Multi-Agent Planning}
\noindent\textbf{Bimanual Manipulation.}
With the development of deep learning techniques, some progress \cite{Drolet2024} has been made in learning dexterous manipulation skills using two grippers \cite{Grannen2023} or human-like hands \cite{chen2022towards}, like scooping food \cite{Grannen2022}, folding clothes \cite{avigal2022speedfolding}, zipping zippers \cite{Grannen2023}, etc. However, these works focus on designing learning-based systems to perform human-level operation skills, neglecting the explicit planning abilities in complex long-horizon tasks. 
As illustrated in these works, the high complexity associated with the variety of bimanual patterns suggests that high-level planning should be considered as well for an integrated control system design \cite{Krebs2022, Smith2012}.

\noindent\textbf{Bimanual and Multi-Agent Planning.}
Early works explored using symbolic methods for designing multi-agent systems to generate an efficient plan for complex tasks in a textual world \cite{torreno2014fmap, cox2009efficient, ghallab2004automated}. Recent works have leveraged LLMs for multi-agent collaboration, where either a centralized or distributed LLM analyzes, decomposes, and assigns tasks to agent candidates in both textual \cite{Hong24MetaGPTMeta} and robotic environments \cite{Kannan24SMARTLLMSmart}.
A bimanual robot can be viewed as a specialized multi-agent system, typically comprising homogeneous agents (arms) operating within a confined workspace (space around the robot body). Unlike many multi-agent systems where robots handle subtasks with varying degrees of independence, bimanual planning involves tightly coupled interactions between two arms, making it uniquely constrained in both spatial and temporal dimensions.
For bimanual task planning, recent work DAG-Plan \cite{gao2024dag} employs LLMs to decompose tasks into directed acyclic graphs, assigning them to the left and right arms based on predefined rules that account for availability. However, existing approaches heavily depend on the reasoning and coding capabilities of LLMs for accurate planning, overlooking the need for computational search in an abstract space to optimize coordination. Additionally, integrating abstract representations could enhance self-verification before execution.

To obtain robust and efficient bimanual plans, we format the spatial scene information for the task with linguistic task descriptions into PDDL definitions, enabling a partial-order plan to be generated for bimanual tasks, with guaranteed logical correctness and higher efficiency.

\begin{figure}[t!]
 \begin{center}
  \centerline{\includegraphics[width=0.8\columnwidth]{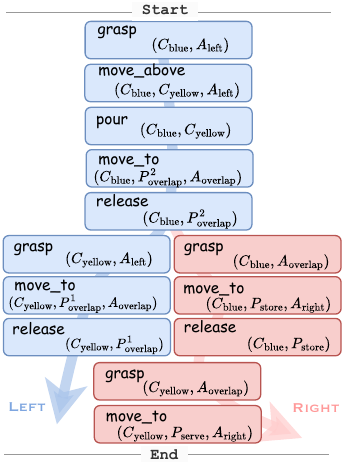}}
 \end{center}
    \vskip -1.5em
     \caption{Illustrative partial-order plan for bimanual manipulation (cf. Section~\ref{sec:method}), where $C$, $A$ and $P$ indicate \textit{cup}, \textit{area} and \textit{point} respectively. Actions for the left and right hands are colored in light blue and red respectively.  Two boxes shown horizontally side by side represent two actions executed in parallel.}
    \label{fig:partial_order_plan}
\end{figure}

\section{Method}\label{sec:method}

\begin{figure*}[ht!]
 \begin{center}
  \centerline{\includegraphics[width=1\textwidth]{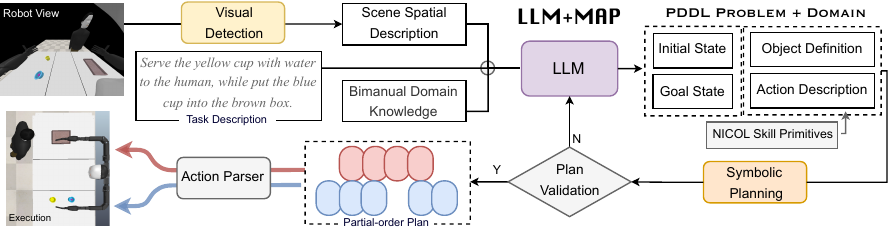}}
 \end{center}
     \caption{Overview of our framework. According to the spatial description of the scene, with the bimanual domain knowledge and task description, LLM+MAP generates a PDDL representation that is used for multi-agent symbolic planning. Then, a valid partial-order plan is generated and executed by the NICOL bimanual robot (see Figure~\ref{fig:three_images} for scenario setting).}
    \label{fig:method}
\end{figure*}

Given a task description in natural language, we aim to generate a valid and efficient plan for a bimanual robot based on the initial task configurations, e.g., spatial information about the objects with respect to the robot. In this sense, this section introduces the three elements of LLM+MAP: spatial scene description, PDDL definitions with a multi-agent solver for the bimanual robotic scenario, and the LLM as a PDDL writer.

\subsection{Spatial Scene Understanding in Bimanual Robots}
Based on spatio-temporal control patterns introduced in previous work \cite{Chu2024}, we define the manipulation areas for the hands based on the reachable areas:
\begin{itemize}
    \item Uncoordinated Areas: areas that only one hand can reach, whereas the other one cannot, i.e., the left area and right area respectively. In those two areas, the two hands can act independently and do not influence each other; thus, manipulations can naturally occur in parallel.
    \item Coordinated Area: the area that both hands can reach, i.e., the overlap area in the middle. The two hands act and manipulate dependently in spatial and temporal relations. They can collaborate either asynchronously or synchronously -- an asynchronous type of control involves one hand constructing pre-conditions for the other, while a synchronous control indicates a, usually precise, mutual dependency between them.
\end{itemize}
Based on this operational paradigm, for a bimanual task, we first need to figure out the areas in which the task-relevant objects and positions are located. To do this, we marked black lines on the edges of the work area to visualize and distinguish different areas. When receiving a task description, we use object-detection models like OWLv2 \cite{minderer2023scaling} to locate objects on the desktop according to the task-related text queries. Then, based on the object bounding box information compared with the black lines on the image, a rule-based recognizer is used to determine the area in which each object is located.

\subsection{Partial-order Plan Generation with LLMs and PDDL}
We formalize the bimanual manipulation scenario as a special case of multi-agent task planning problems. In this section, we will first informally introduce the preliminaries of PDDL, and then define the bimanual domain in PDDL. We refer the reader to other references \cite{kovacs2012multi, torreno2014fmap} for a formal treatment.

\subsubsection{Planning Domain Definition Language}
Planning Domain Definition Language (PDDL, \cite{aeronautiques1998pddl}) is a declarative language for standardizing the formalization of planning problems. In general, a planning problem is characterized as a tuple $\mathcal{T}$: $$\mathcal{T} = <\mathcal{O}, \mathcal{P}, \mathcal{A}, \mathcal{S}, \mathcal{G}>,$$
where $\mathcal{O}$ is the set of objects, $\mathcal{P}$ is the set of predicates, $\mathcal{A}$ is the set of actions, $\mathcal{S}$ is the initial state, and $\mathcal{G}$ is the goal state. A state is defined as a list of predicates applied to objects and agents, that hold. Each action is defined with parameters specifying input types and describes the pre-conditions and effects of executing such action, in terms of a series of predicates for the object inputs. The PDDL presentation of a planning problem consists of two files, 1) a \textit{domain}, which defines objects, predicates, and actions with pre-condition and effect specifications that describe the task world, and 2) a \textit{problem}, which includes an initial state and a desired goal state which are the sets of several specific predicates. 

\subsubsection{Multi-agent Planning}
In multi-agent problems, there can be a set of agents from the set of agents, $\mathcal{R}=\left\{R_{1}, ..., R_{N}\right\}$. For an agent $R_{i}\in \mathcal{R}$, it can have its own set of actions $\mathcal{A}^{i}$, predicates $\mathcal{P}^{i}$, initi states $\mathcal{S}^{i}$, and goal states $\mathcal{G}^{i}$. In the cooperative multi-agent problems, the goal of each agent remains the same, i.e., $\mathcal{G}^{i}=\mathcal{G}$, and only by cooperating can the goal be accomplished. In $\mathcal{P}^{i}$, there are public predicates from $\mathcal{P}$ and a certain number of private predicates, i.e., environment-common and agent-specific properties. When executing an action inside one's own action set, one needs to consider if the value of the predicate from $\mathcal{P}^{i}$ and the environment's predicate $\mathcal{P}$ meet the action's pre-conditions. Besides, a cooperative action among agent $i$ and $j$ can only be executed if the value for several predicates from $\mathcal{P}^{i}$, $\mathcal{P}^{j}$ and $\mathcal{P}$ are hold. In this sense, the common goal can be distributed to different agents and accomplished through their cooperation via interactions. To this end, the plan generated is a partial-order plan, which is a sequence of actions with dependencies but also flexibilities in execution orders, enabling parallel processes in multi-agent systems \cite{ghallab2004automated}. A dominant method for such type of problems is FMAP (Forward MultiAgent Planning, \cite{torreno2014fmap}), which effectively handles both cooperative and independent planning problems, outperforming existing multi-agent planning systems.

Inspired by the above works, we formalize a bimanual robot to a multi-agent system, with \textsc{Left} and \textsc{Right} as two separate agents. With private predicate \textit{control}, each of them has their exclusive right to control the left and right hand respectively. With a cooperative goal, they will have operations in their areas, and an overlap area in between for interactions. 

\subsubsection{PDDL for Bimanual Robotic Scenario}
In the bimanual robotic scenario, to specify the spatio-temporal patterns we introduced and fundamental logics about manipulations, we design a list of predicates, shown in Table \ref{tab:pddl_def}. We define object types in $\mathcal{O}$ as: \textit{hand}, \textit{area}, \textit{object} and \textit{point}, and define a list of predicates for them, including their properties and relationships between them. Based on these definitions, we can define symbolic actions with parameter inputs and their pre-conditions and effects descriptions. 
We design atomic and fundamental skills for the homogenous robotic hand, catering for both independent (single-hand) and joint operations, shown in Table~\ref{tab:nicol_skills}. For a joint operation involving both hands, the action is not defined with specific hand(s) as input parameters (since only one hand can be controlled through the agent’s private predicate). Instead, the precondition requires both hands to be in an \texttt{available} state. Therefore, joint operations are a unique class of actions executed by both hands, despite being initiated by a single agent.

\begin{table}[ht!]
\centering
    \caption{Definition of PDDL predicates for the bimanual robot scenario. By abstracting \textit{hand}, \textit{area}, and \textit{object}, the predicates capture affordance properties and relationships, enabling solutions that generalize across diverse task domains.}
    \label{tab:pddl_def}
    \fontsize{7pt}{7pt}\selectfont
    \renewcommand{\arraystretch}{1.35} 
\begin{tabular}{ll}
\toprule
\textbf{Properties} & \textbf{Relationships}\\[0pt]
\hline
\texttt{control(hand)} & \texttt{arm\_at(hand, area)}\\
\texttt{available(hand)} & \texttt{arm\_access(hand, area)}\\
\texttt{is\_graspale(object)} & \texttt{lifting(object, hand)}\\
\texttt{is\_free(object)} & \texttt{object\_at\_area(object, area)}\\
\texttt{is\_releasable(point)} & \texttt{object\_at\_point(object, point)}\\
\texttt{is\_accessible(point)} & \texttt{point\_at(point, area)}\\
\bottomrule
\end{tabular}
\end{table}

\begin{table}[ht!]
    \centering
    \caption{Bimanual skills design. Single skills are designed independently for single hands, whereas joint skills are designed for two-hand symmetric manipulations with cooperation. Additional verification parameters are for PDDL definition to maintain appropriate pre-condition verification.}
    \label{tab:nicol_skills}
    \fontsize{7pt}{7pt}\selectfont
    \renewcommand{\arraystretch}{1.35} 
    \begin{tabular}{lll}
        \toprule
        \textbf{Type} & \textbf{Skill(Parameters)} & \textbf{Verif. Param.}\\[0pt]
        \hline
        Single & \texttt{grasp(hand, object)} & \texttt{area}\\
        & \texttt{move\_to(hand, object, point)} & \texttt{area}\\
        & \texttt{release(hand, object)} & \texttt{point, area}\\
        & \texttt{push(hand, object, target\_area)} & \texttt{source\_area}\\
        & \texttt{pour(hand)} & \texttt{object1,object2} \\ 
        & \texttt{move\_above(hand, source\_object, } & \texttt{area}\\
        &  \:\:\:\:\:\:\:\:\:\:\:\:\:\:\:\:\texttt{target\_object)} &\\
        & \texttt{place(hand, source\_object, } & \texttt{area}\\
        &  \:\:\:\:\:\:\:\:\:\:\:\:\:\:\:\:\texttt{target\_object)} &\\
        Joint & \texttt{co\_hold(object)} & - \\
        & \texttt{co\_move\_to(point)} & - \\
        \bottomrule
    \end{tabular}
\end{table}

\begin{figure*}[!ht]
    \centering
    \begin{subfigure}{0.32\textwidth}
        \centering
        \includegraphics[width=\linewidth]{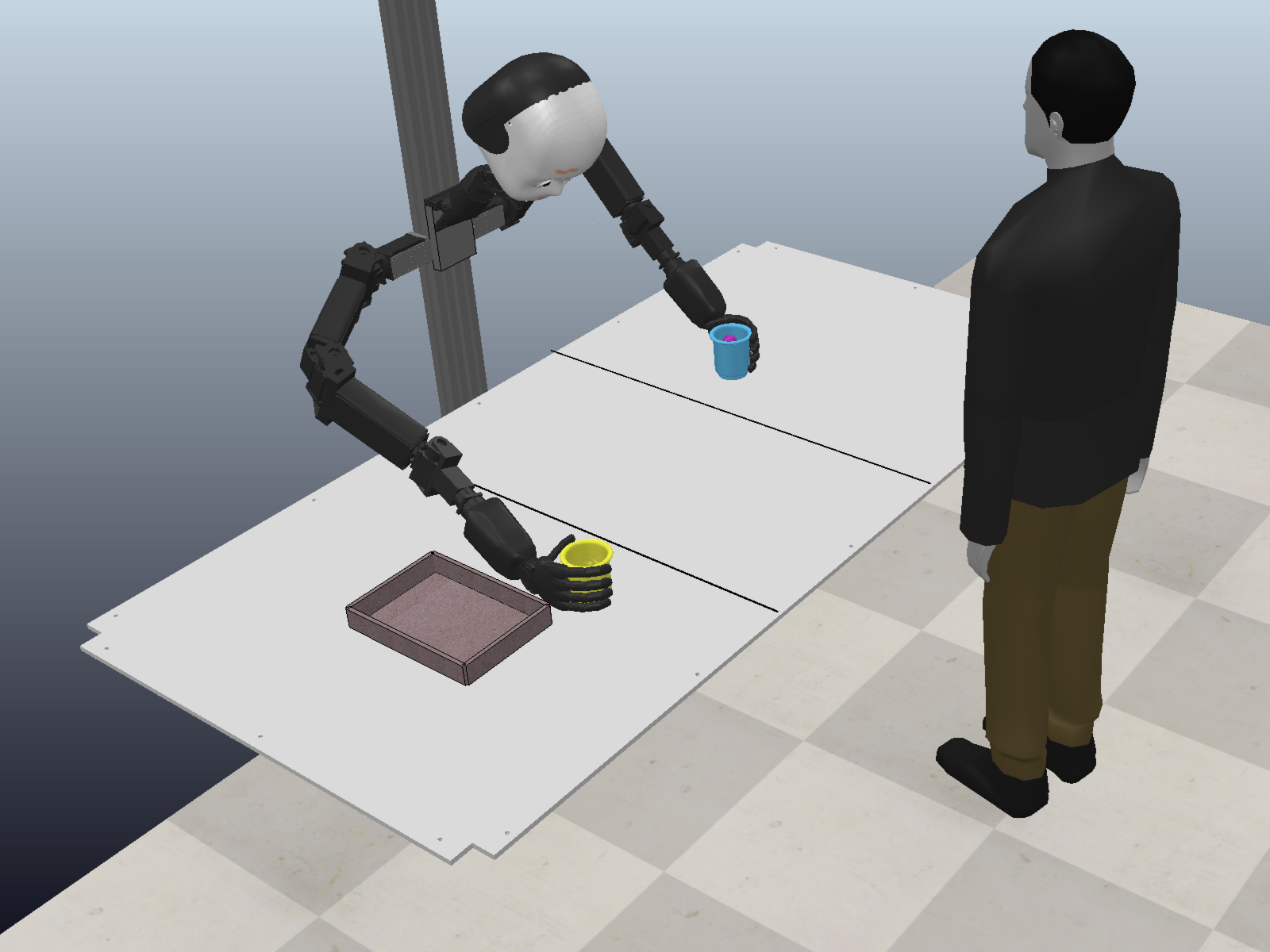}
        \caption{ServeWater}
    \end{subfigure}
    \hfill
    \begin{subfigure}{0.32\textwidth}
        \centering
        \includegraphics[width=\linewidth]{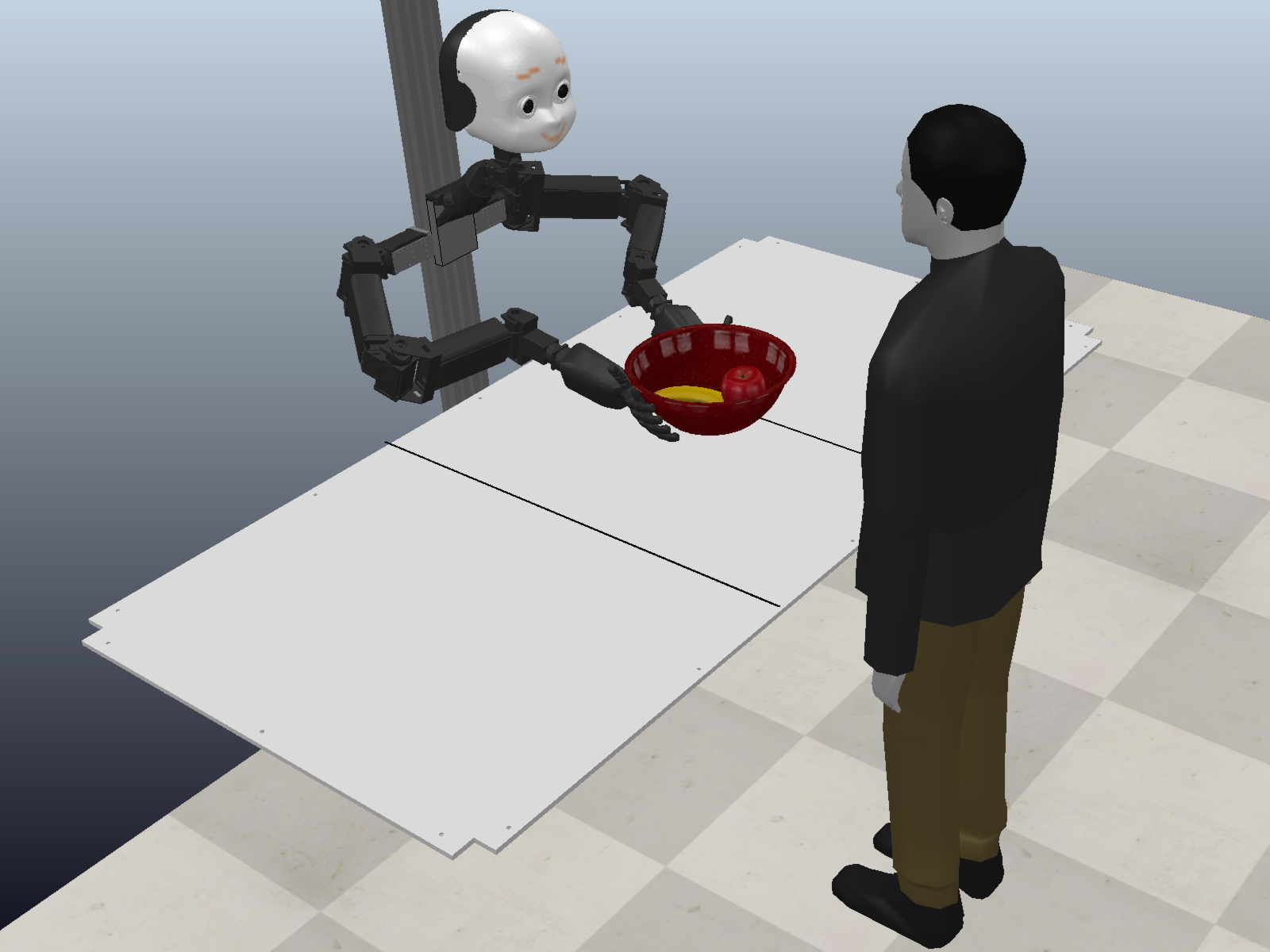}
        \caption{ServeFruit}
    \end{subfigure}
    \hfill
    \begin{subfigure}{0.32\textwidth}
        \centering
        \includegraphics[width=\linewidth]{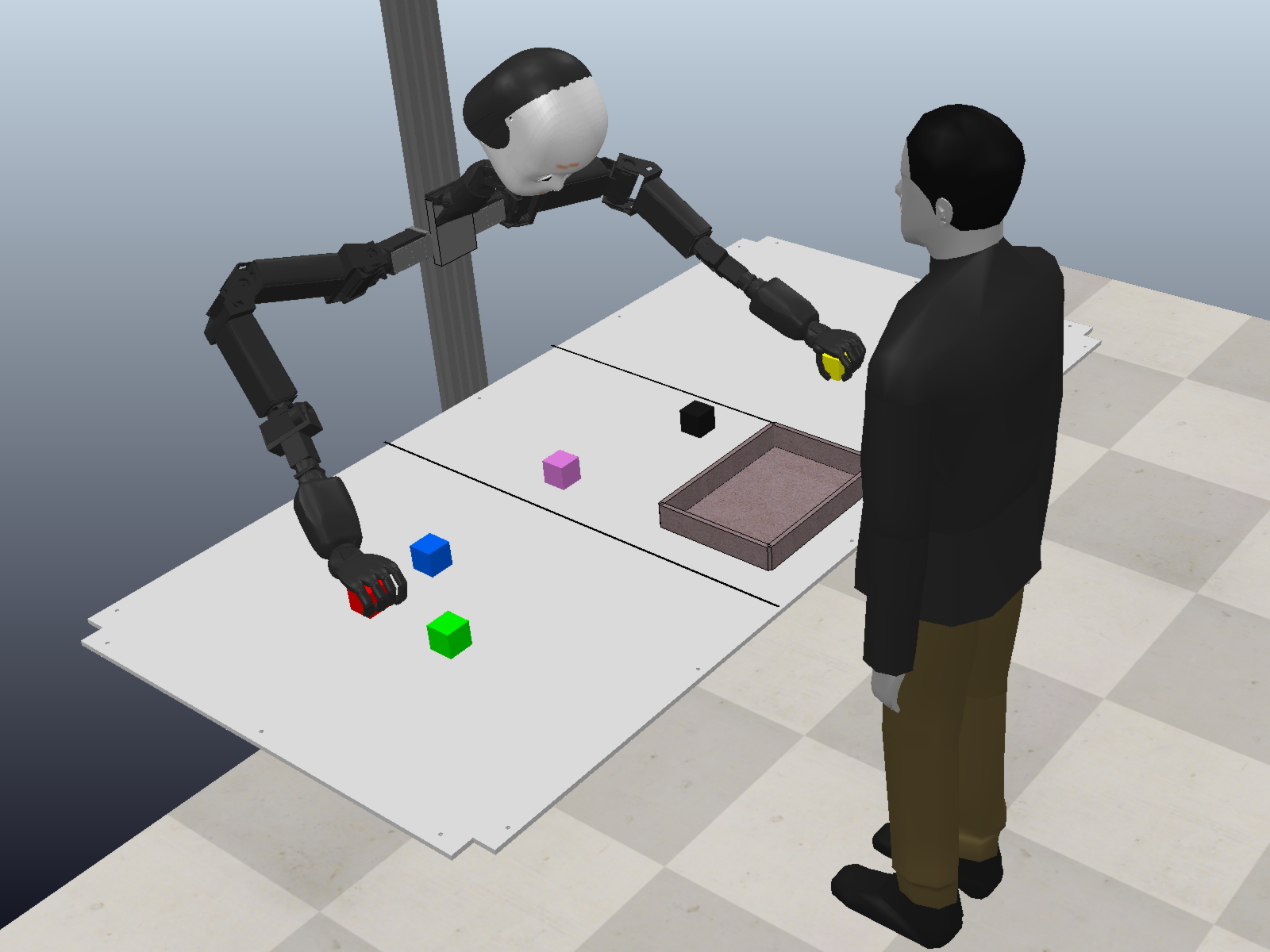}
        \caption{StackBlock}
    \end{subfigure}
    \caption{A visualization of the three task domains. In ServeWateer, the brown box is placed either in the left or right area to store the blue cup, while the cups and the human user are in random areas. In ServeFruit, the human stands exclusively in front of the overlap area to receive the bowl, while the fruits and the bowl are in random areas. In StackBlock, the blocks are distributed at random positions over the three areas, while the human user stands in front of a random area.}
    \label{fig:three_images}
\end{figure*}

\subsubsection{LLMs as a PDDL Writer} Under the above PPDL definition principles, we prompt the LLM to generate the initial state definition according to scene descriptions from the camera image, and the goal state definition based on the input of natural language task descriptions. Since LLMs can effectively understand and generate Python code with minimal errors, we utilize the Unified-Planning Python library \cite{unified_planning_2025}, which streamlines problem definition and simplifies the invocation of built-in symbolic solvers, including the well-known Fast-Downward \cite{helmert2006fast} and FMAP (\cite{torreno2014fmap}) solvers. To reduce coding errors generated by the LLM, we re-prompt it with error messages from the Python interpreter when an error occurs, enabling iterative improvements through regeneration.

\section{Experiments}
We conduct the following experiments to evaluate to what extent our method can achieve efficient bimanual control in robotic tasks. Specifically, we evaluate the success and efficiency of the proposed LLM+MAP in three task domains. 
We built LLM+MAP on GPT-4o as the backend. As a baseline, GPT-4o is prompted with bimanual domain knowledge to generate executable plans directly. The complexity of the task necessitates stronger reasoning ability for direct plan generation. Therefore, given recent advancements in reasoning-capable LLMs, we additionally conduct experiments with o1 and R1 model as two strong baselines. 
Besides, we include the V3 model as a counterpart to R1, as it has the same size but lacks reasoning capabilities.

\subsection{Environment Setup}

The experiments are conducted in the CoppeliaSim\footnote{https://www.coppeliarobotics.com/} simulator\footnote{In the simulation, we do not incorporate physical randomness (e.g., certain probability of failure) to mimic real-world execution, as our primary focus is on the planning domain. Introducing physical randomness would add additional uncertainty, which could complicate the analysis.} on the NICOL robot \cite{Matthias2023, Zhao2023}, which is designed to blend social interaction with reliable object manipulation capabilities. 
We use one of the two cameras at its eye positions for visual perception. At the manipulation level, it is equipped with two arms, each with six degrees of freedom, and an adult-sized, five-fingered manipulator attached to it for precise manipulation of everyday objects. 

Three bimanual task domains are designed in a human-robot environment: ServeWater, ServeFruit, and StackBlock. Such a setting closely resembles the real-world interaction, and also brings about the complexity and diversity of tasks through the spatial relationship between the human and the task-related objects.

In the ServeWater task domain, there are two cups on the work table, an empty yellow cup and a water-filled blue cup, and the task requires the robot to serve the water in the yellow cup to the human, while putting the blue cup to a specified store point on the brown box, which is randomly located in the left or right area. In the ServeFruit task domain, there is a banana, an apple, and a big red bowl, and the task requires serving the fruits with a bowl to the human. It should be noted that the bowl is not graspable with one single hand, and the symmetric manipulations for two hands can only be done in the overlap area. In this sense, the bowl should be pushed to the overlap area so that it can be held by both hands. In the StackBlock task domain, there are several blocks with different colors randomly located on the table, and the task goal is to stack one or two piles with specific blocks in a certain order. We design the task's configurations in two aspects: the total number of blocks to be stacked, and the number of cubes in each pile. Specifically, the task requires the robot to stack the selected four or five blocks into a specific pile in terms of $[(2,2),(3,1),(4,0)]$ or $[(2,3),(4,1),(5,0)]$, respectively. An example of a task goal with $(4,1)$ in natural language is, the task is to stack the selected five blocks in two piles on the tray right in front of the human, where one pile is yellow over red over purple over black and one pile is green.

\subsection{Metrics}\label{subsec:metrics}
To examine the efficacy and efficiency of generated plans, we compare our proposed methods with baselines along the following three metrics:
\begin{itemize}
    \item Planning Time (PT). For LLM-direct task planning, the time cost depends solely on the inference time of certain LLMs, while the planning time of our method is composed of LLM inference time and symbolic planning time.
    \item Success Rate (SR), which reveals the overall ability for task completion.
    \item Group Debits (GD), a metric to compare the planning efficiency of models.\footnote{Since the minimum number of task completion steps varies across runs, due of the randomness in initialization, a comparative metric avoids the need to know the number of steps of the optimal plan.} 
    Specifically, we set the debits of the ``champion" model (the one that uses the fewest planning steps) to 0. Other models are assigned debits according to the number of planning steps exceeding those of the champion (higher debits mean worse performance). 

\end{itemize}


\subsection{Results}

\noindent{\textbf{Planning Time.}} As is shown in Table~\ref{tab:plan_time}, we have several immediate observations: 
(1) the PT of GPT-4o, V3 direct and LLM+MAP are dramatically shorter than the other two advanced reasoning models, o1 and R1.
(2) The LLM inference time increases across all models as task complexity grows.
(3) In analyzing the PT distribution for LLM+MAP, the LLM inference time remains relatively stable, while the symbolic planning time for simpler tasks is sufficiently small (smaller than the LLM inference time), but increases considerably for more complex tasks with a larger search space.
\begin{table}[ht]
\centering
    \caption{Task Planning Time ($s$) for different models. For LLM+MAP, we also provide the time spent on each module, including LLM inference time for code generation and PDDL planning time.}
    \label{tab:plan_time}
    \fontsize{7pt}{7pt}\selectfont
    \renewcommand{\arraystretch}{1.5} 
\begin{tabular}{lcccc}
\toprule
\textbf{Model} & \textbf{ServeWater} & \textbf{ServeFruit} & \textbf{StackBlock-4} & \textbf{StackBlock-5}\\[0pt]
\hline
gpt-4o direct & 5.33 \tiny{±6.70} & 4.17 \tiny{±1.35} & 7.82 \tiny{±2.46} & 11.17 \tiny{±20.65}\\[0pt]
V3 direct & 10.80 \tiny{±1.90} & 10.41 \tiny{±2.03} & 14.00 \tiny{±2.41} & 17.20 \tiny{±2.78}\\[0pt]
R1 direct & 122.42 \tiny{±40.16} & 144.42 \tiny{±70.94} & 220.85 \tiny{±72.68} & 168.28 \tiny{±38.59}\\[0pt]
o1 direct & 104.29 \tiny{±71.69} & 77.68 \tiny{±31.62} & 96.85 \tiny{±51.00} & 66.38 \tiny{±17.96}\\[0pt]
\hline
LLM+MAP & 11.34 \tiny{±6.73} & 7.75 \tiny{±0.91} & 34.97 ± \tiny{21.53}& 62.75 \tiny{±26.05} \\[0pt]
\:\: $\in$ LLM & 9.55 \tiny{±6.70} & 6.00 \tiny{±0.91} & 13.08 \tiny{±4.19} & 18.23 \tiny{±8.11}\\[0pt]
\:\: $\in$ MAP & 1.80 \tiny{±0.49} & 1.74 \tiny{±0.25} & 21.89 \tiny{±20.97} & 44.52 \tiny{±23.40}\\[0pt]
\bottomrule
\end{tabular}
\end{table}
In practice, multi-agent planning can be time-consuming with the FMAP solver, so we set a \textit{timeout} to avoid excessively long computation times. As an alternative, we convert it to a single-robot task, allowing for a feasible solution with the BFWS solver \cite{frances2018best}. According to the bimanual characteristics, the generated plan is then post-processed as a partial-order plan using automated graph tools.

\noindent\textbf{Success Rate.} From the experimental results presented in Table~\ref{tab:success_rate}, we have the following findings:
(1) It is clear that our proposed method achieves the highest performance out of all tasks, thanks to the integration of LLM coding and multi-agent planning.
This result is impressive because the original performance of GPT-4o, as a base LLM with specifically tuning in the pursuit of strong reasoning ability, performs poorly while our method -- built on top of this base model -- outperforms. GPT-4o's strength in in-context understanding minimizes errors during the generation of PDDL definitions. In the StackBlock domain, such errors primarily result from GPT-4o defining the task goal with an incorrect block order, which consequently leads to execution failures in the actual task configuration.
(2) Long-horizon robotic tasks require strong reasoning ability for correct task completions, specifically tuned reasoning LLMs, o1 and R1, trade inference time compute (indicating both higher cost in both time and expense) with a better reasoning competence, resulting in higher SR in our experiments.
(3) Comparing the results of V3 and R1, it is enlightening that, \textit{despite having the same parameter scale}, R1 dramatically outperforms V3. This suggests that strong reasoning ability is crucial for solving long-horizon tasks, particularly when the base language model is not integrated with a planning mechanism like LLM+MAP.

\begin{table}[h]
    \caption{Success rate (\%) of plan execution across the task domains.} 
    \label{tab:success_rate}
    \centering
    \renewcommand{\arraystretch}{1.2} 
    \begin{tabular}{lcccc}
        \toprule
        \scriptsize{\textbf{Model}} & \scriptsize{\textbf{ServeWater}} & \scriptsize{\textbf{ServeFruit}} & \scriptsize{\textbf{StackBlock-4}} & \scriptsize{\textbf{StackBlock-5}} \\ 
        \hline
        gpt-4o direct& 2 & 13 & 2 & 0 \\ 
        V3 direct & 2 & 6 & 6 & 1 \\
        R1 direct & 67 & 63 & 94 & 77 \\ 
        o1 direct & 84 & 82 & 95 & 88 \\ 
        LLM+MAP & \textbf{100} & \textbf{100} & \textbf{96} & \textbf{97}\\ 
        \bottomrule
    \end{tabular}
\end{table}

\begin{figure*}[!h]  
    \centering
    \begin{subfigure}{0.24\textwidth}
        \centering
        \includegraphics[width=\linewidth]{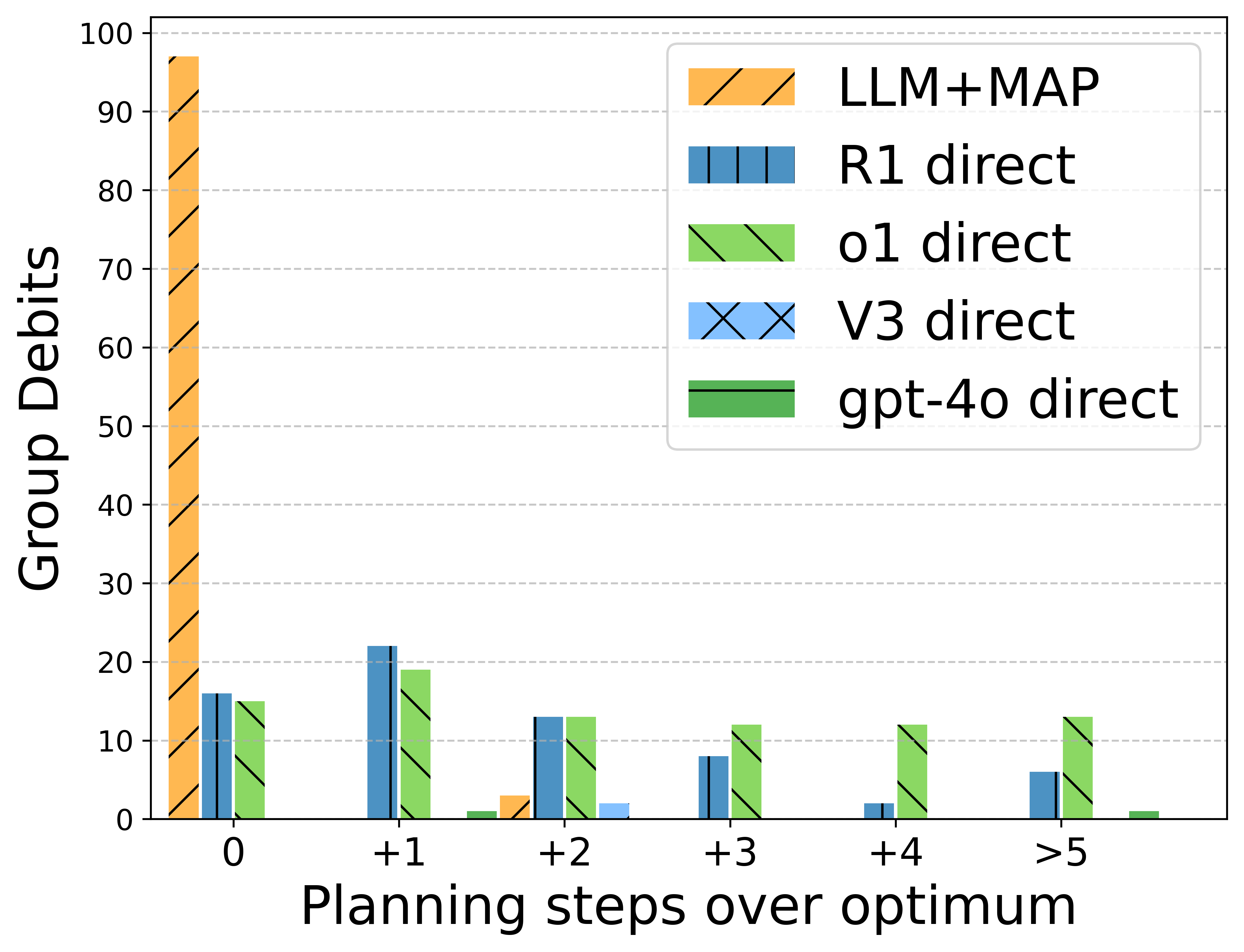}
        \caption{ServeWater}
    \end{subfigure}
    \begin{subfigure}{0.24\textwidth}
        \centering
        \includegraphics[width=\linewidth]{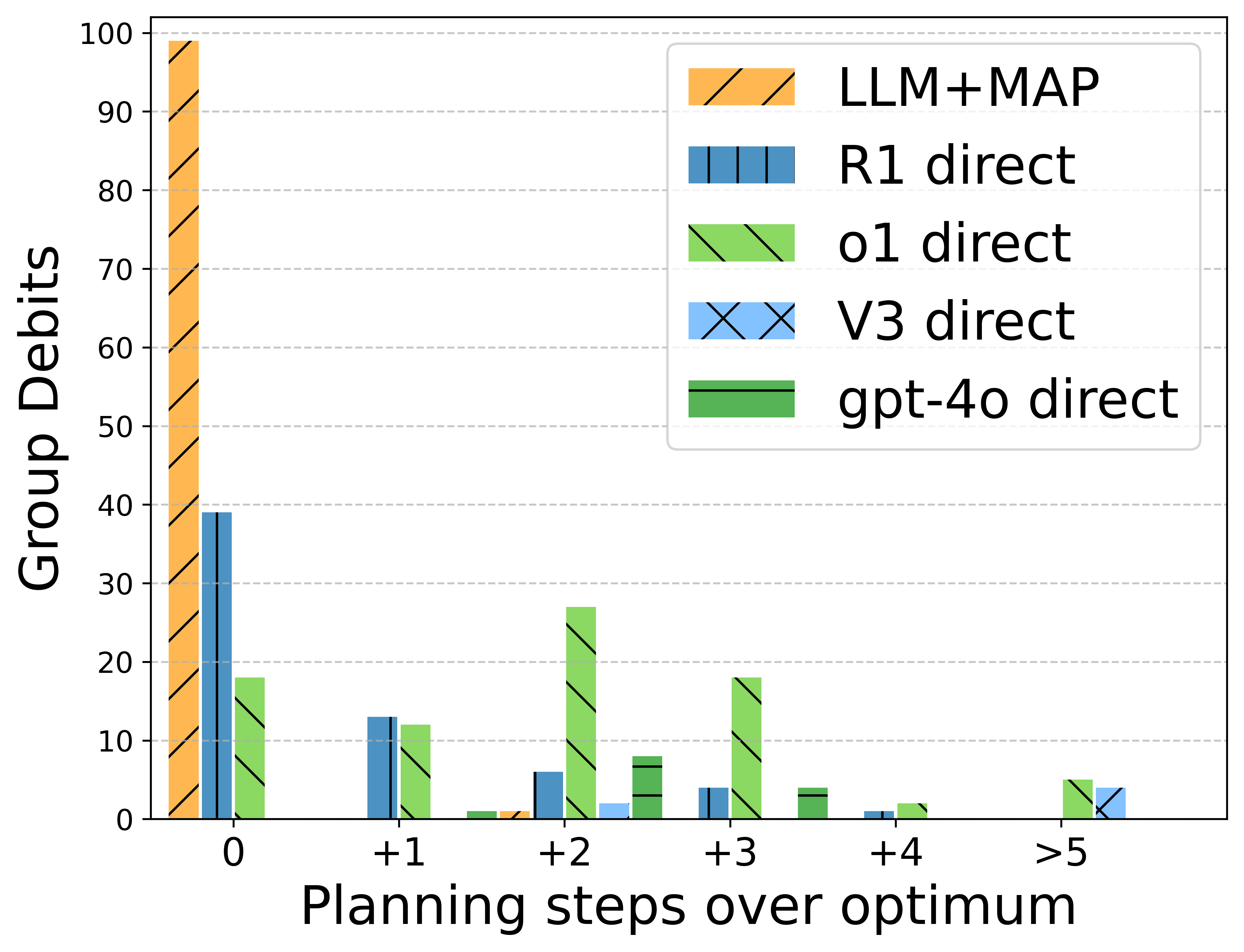}
        \caption{ServeFruit}
    \end{subfigure} 
    \begin{subfigure}{0.24\textwidth}
        \centering
        \includegraphics[width=\linewidth]{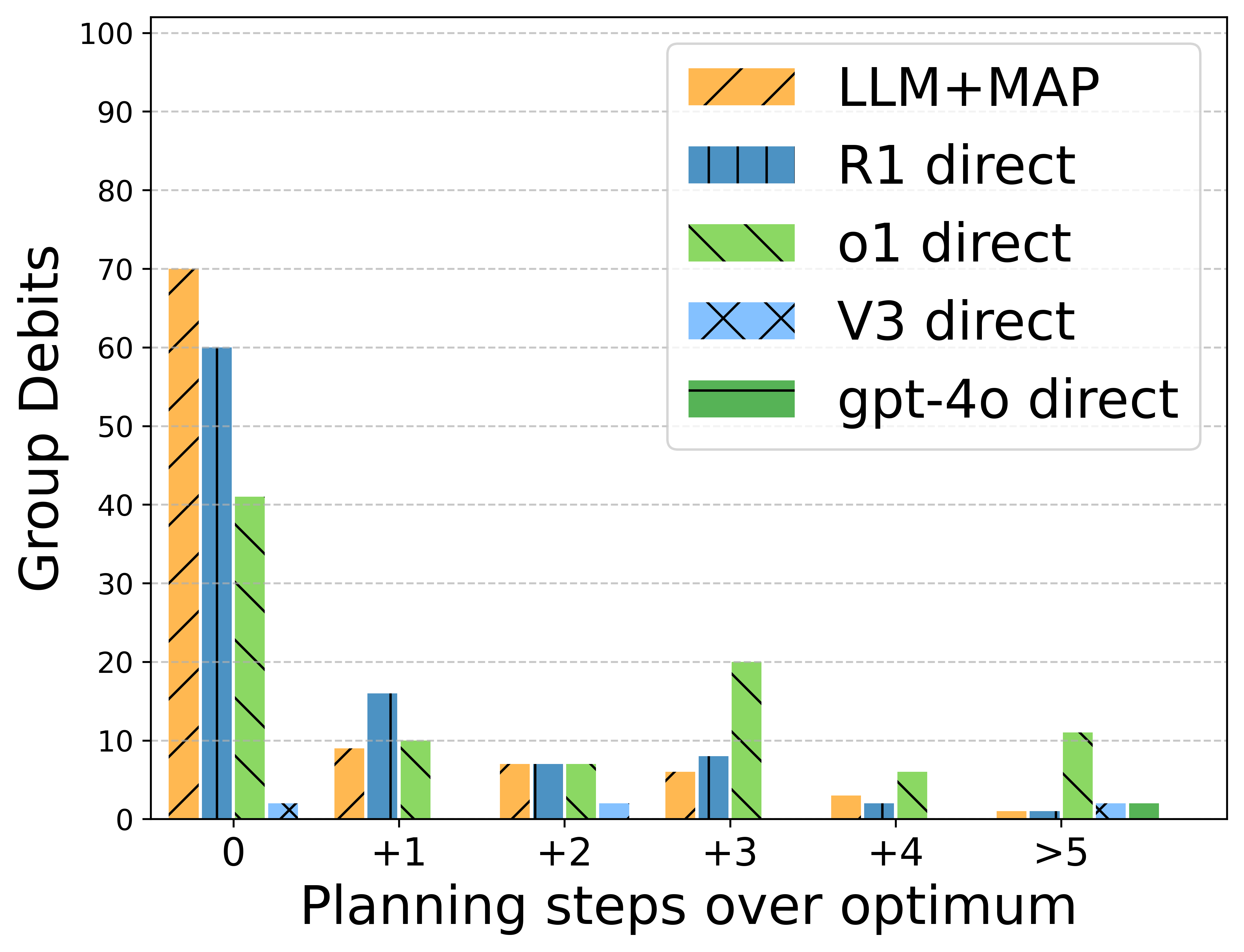}
        \caption{StackBlock-4}
    \end{subfigure}
    \begin{subfigure}{0.24\textwidth}
        \centering
        \includegraphics[width=\linewidth]{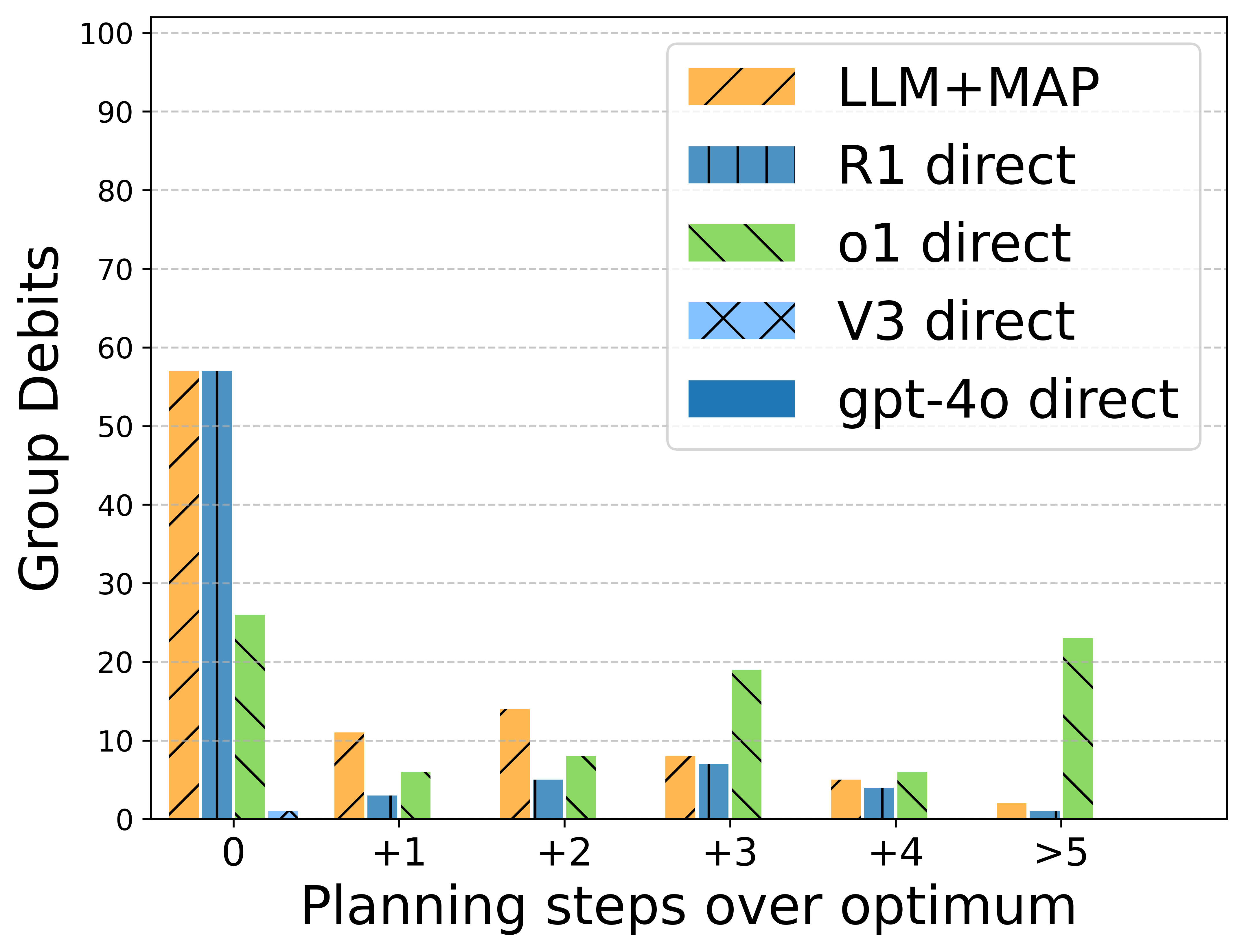}
        \caption{StackBlock-5}
    \end{subfigure}
    \caption{The \textit{Group Debits} statistics among successful tasks in three domains, the smaller the better.}
    \label{fig:relative_plan_steps}
\end{figure*}

\noindent{\textbf{Group Debits.}} As is discussed in Subsection~\ref{subsec:metrics}, GD is a metric to compare the planning
efficiency of models. From Figure~\ref{fig:relative_plan_steps}, we find that 
(1) LLM+MAP almost dominates the competitions, especially in easier tasks, with mass mainly distributed around 0 debits (relative optimal, i.e. winner with minimal planning steps among competitors).
(2) With the growth of task complexity, the GD of reasoning models is comparable to the multi-agent planning results, indicating that stronger reasoning ability helps a comprehensive understanding of the temporal and spatial resilience of dual hands. However, for easier tasks, plans generated by those strong reasoning models are far from being efficient, we hypothesize that this non-efficiency may stem from the overthink \cite{chen2024not} of current reasoning models, especially of the R1 model, which takes excessive reasoning time (cf. Table~\ref{tab:plan_time}) and outputs wordy content.

\noindent\textbf{Discussion.} 
The combination of the above metrics provides additional insights. Although the SR of the R1 model is lower than that of the o1 model, its GD scores higher, indicating greater efficiency in terms of bimanual coordination and suggesting a superior temporal understanding. In contrast, the o1 model appears to have a stronger spatial understanding, which is crucial for successful task completion. The decoupling of temporal and spatial dimensions, along with the corresponding investigation, is left as a direction for future work.

\subsection{Ablation Study}
\begin{figure}[!th]  
    \centering
    \includegraphics[width=\linewidth]{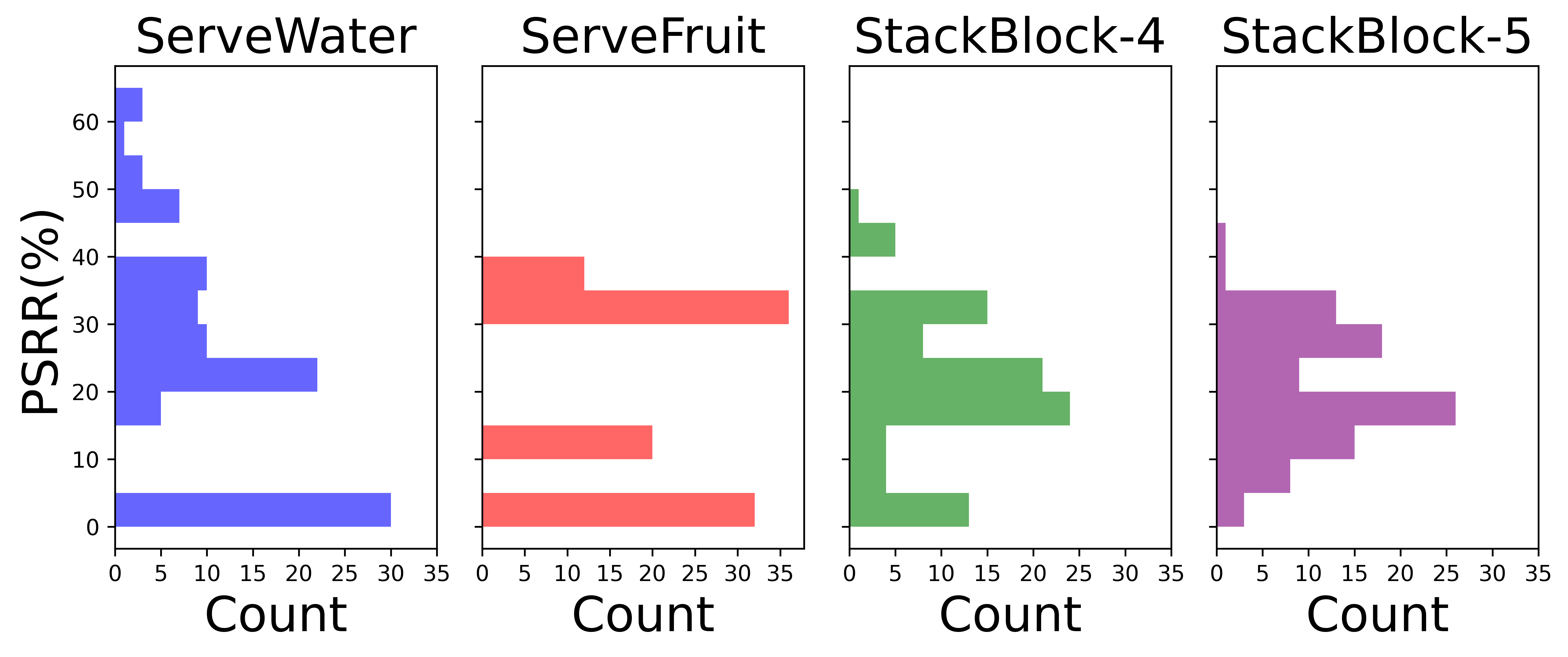}
    \setlength{\abovecaptionskip}{-10pt}
    \caption{Planning Step Reduction Rate (\%) of LLM+MAP over LLM+P, showcasing the improved efficiency of Multi-agent planning.} 
    \label{fig:ablation_PSRR}
\end{figure}

To investigate the efficacy of multi-agent PDDL planning in comparison to traditional planning, which treats configurations of both hands as a unified collection of presets of a single agent, we conduct an ablation study by removing MAP component and only use BFWS \cite{frances2018best} as its solver to generate sequential plan, resulting in an adapted\footnote{Note that the method introduced in LLM+P never considers dual robotic arm or other forms of multi-agent systems, we thus adapt it onto bimanual setting with modification ranging from skill design to domain definition. Implementing details can be found in the code page.} implementation of LLM+P \cite{liu2023}. 
We compute the Planning Step Reduction Rate (PSRR) as $$\text{PSRR} = \frac{N_\text{+P} - N_\text{+MAP}}{N_\text{+P}} \times 100\%, $$ where $N_\text{+P}$ and $N_\text{+MAP}$ are the number of planning steps for LLM+P and LLM+MAP respectively.
As shown in Figure~\ref{fig:ablation_PSRR} (across successful tasks in 100 runs), compared to the sequential plans generated by LLM+P, our method facilitates more parallel task allocation for both hands, resulting in a significant improvement in overall efficiency.

\section{Conclusion}
We propose LLM+MAP, which formulates the bimanual robot task planning problem as a special form of multi-agent planning, leveraging the coding and reasoning ability of LLMs and formalized language representations for efficient multi-agent planning. Specifically, vision models transform spatial information and task descriptions into PDDL representations in bimanual robot scenarios, achieving efficient spatial and temporal coordination with guaranteed correctness and optimality by symbolic planners. Extensive experiments on three task domains showcase that our framework outperforms planning results with GPT-4o, V3 and even strong reasoning models o1 and R1, in terms of higher success rate and efficiency with less plan generation time. 

\noindent\textbf{Limitations and Future Work.} While our main focus in this paper is not on the acquisition and design of bimanual skills, the execution of the underlying task assumes that existing motion primitives serve the purpose well, raising a certain gap when transferring to the real world. Integrating with learning-based bimanual robotic skills will be our main focus in the future. We have used the LLM+MAP framework for merely two agents, i.e., the robot's hands, while the feasibility and effectivity of extending our framework to control a larger number of agents are still to be investigated. Hierarchical planning with LLM bootstrapping can be a viable direction to explore for task planning in large-scale multi-robot tasks. Additionally, considering the dynamic nature of the world, future work includes incorporating action durations or costs for more fine-grained bimanual planning, enabling adaptive re-planning in response to action failures or unexpected environmental changes, and improving the verification of action preconditions and effects, etc.

\section*{Acknowledgment}
The authors gratefully acknowledge support from the Horizon Europe project TERAIS and the MSCA Doctoral Network TRAIL. Additionally, they express their gratitude to OpenAI's Researcher Access Program for generously providing API tokens.

\addtolength{\textheight}{-1cm}
\bibliographystyle{IEEEtran}
\bibliography{references}

\begin{thebibliography}{10}
\providecommand{\url}[1]{#1}
\csname url@rmstyle\endcsname
\providecommand{\newblock}{\relax}
\providecommand{\bibinfo}[2]{#2}
\providecommand\BIBentrySTDinterwordspacing{\spaceskip=0pt\relax}
\providecommand\BIBentryALTinterwordstretchfactor{4}
\providecommand\BIBentryALTinterwordspacing{\spaceskip=\fontdimen2\font plus
\BIBentryALTinterwordstretchfactor\fontdimen3\font minus \fontdimen4\font\relax}
\providecommand\BIBforeignlanguage[2]{{%
\expandafter\ifx\csname l@#1\endcsname\relax
\typeout{** WARNING: IEEEtran.bst: No hyphenation pattern has been}%
\typeout{** loaded for the language `#1'. Using the pattern for}%
\typeout{** the default language instead.}%
\else
\language=\csname l@#1\endcsname
\fi
#2}}

\bibitem{Krebs2022}
F.~Krebs and T.~Asfour, ``A bimanual manipulation taxonomy,'' \emph{IEEE Robotics and Automation Letters}, vol.~7, no.~4, pp. 11\,031--11\,038, 2022.

\bibitem{Drolet2024}
M.~Drolet, S.~Stepputtis, S.~Kailas, A.~Jain, J.~Peters, S.~Schaal, and H.~B. Amor, ``A comparison of imitation learning algorithms for bimanual manipulation,'' \emph{IEEE Robotics and Automation Letters}, vol.~9, no.~10, pp. 8579--8586, 2024.

\bibitem{shawbimanual}
K.~Shaw, Y.~Li, J.~Yang, M.~K. Srirama, R.~Liu, H.~Xiong, R.~Mendonca, and D.~Pathak, ``Bimanual dexterity for complex tasks,'' in \emph{The 8th Annual Conference on Robot Learning (CoRL)}, 2024.

\bibitem{Grannen2023}
J.~Grannen, Y.~Wu, B.~Vu, and D.~Sadigh, ``Stabilize to act: Learning to coordinate for bimanual manipulation,'' in \emph{The 7th Conference on Robot Learning (CoRL)}.\hskip 1em plus 0.5em minus 0.4em\relax PMLR, 2023, pp. 563--576.

\bibitem{Grannen2022}
\BIBentryALTinterwordspacing
J.~Grannen, Y.~Wu, S.~Belkhale, and D.~Sadigh, ``Learning bimanual scooping policies for food acquisition,'' in \emph{Proceedings of The 6th Conference on Robot Learning}, ser. Proceedings of Machine Learning Research, K.~Liu, D.~Kulic, and J.~Ichnowski, Eds., vol. 205.\hskip 1em plus 0.5em minus 0.4em\relax PMLR, 14--18 Dec 2023, pp. 1510--1519. [Online]. Available: \url{https://proceedings.mlr.press/v205/grannen23a.html}
\BIBentrySTDinterwordspacing

\bibitem{zhao2023survey}
W.~X. Zhao, K.~Zhou, J.~Li, T.~Tang, X.~Wang, Y.~Hou, Y.~Min, B.~Zhang, J.~Zhang, Z.~Dong, \emph{et~al.}, ``A survey of large language models,'' \emph{arXiv preprint arXiv:2303.18223}, vol.~1, no.~2, 2023.

\bibitem{Ahn2022}
A.~Brohan, Y.~Chebotar, C.~Finn, K.~Hausman, A.~Herzog, D.~Ho, J.~Ibarz, A.~Irpan, E.~Jang, R.~Julian, \emph{et~al.}, ``Do as {I} can, not as {I} say: Grounding language in robotic affordances,'' in \emph{Proceedings of The 6th Conference on Robot Learning (CoRL)}.\hskip 1em plus 0.5em minus 0.4em\relax PMLR, 2023, pp. 287--318.

\bibitem{Zhao2023}
X.~Zhao, M.~Li, C.~Weber, M.~B. Hafez, and S.~Wermter, ``Chat with the environment: Interactive multimodal perception using large language models,'' in \emph{2023 IEEE/RSJ International Conference on Intelligent Robots and Systems (IROS)}, 2023, pp. 3590--3596.

\bibitem{Chu2024}
K.~Chu, X.~Zhao, C.~Weber, M.~Li, W.~Lu, and S.~Wermter, ``Large language models for orchestrating bimanual robots,'' in \emph{2024 IEEE-RAS 23rd International Conference on Humanoid Robots (Humanoids)}, 2024, pp. 328--334.

\bibitem{gao2024dag}
Z.~Gao, Y.~Mu, J.~Qu, M.~Hu, L.~Guo, P.~Luo, and Y.~Lu, ``{DAG}-plan: Generating directed acyclic dependency graphs for dual-arm cooperative planning,'' \emph{arXiv preprint arXiv:2406.09953}, 2024.

\bibitem{wang2024survey}
L.~Wang, C.~Ma, X.~Feng, Z.~Zhang, H.~Yang, J.~Zhang, Z.~Chen, J.~Tang, X.~Chen, Y.~Lin, \emph{et~al.}, ``A survey on large language model based autonomous agents,'' \emph{Frontiers of Computer Science}, vol.~18, no.~6, p. 186345, 2024.

\bibitem{liu2023}
B.~Liu, Y.~Jiang, X.~Zhang, Q.~Liu, S.~Zhang, J.~Biswas, and P.~Stone, ``{LLM+P}\: Empowering large language models with optimal planning proficiency,'' \emph{arXiv preprint arXiv:2304.11477}, 2023.

\bibitem{xie2023translating}
Y.~Xie, C.~Yu, T.~Zhu, J.~Bai, Z.~Gong, and H.~Soh, ``Translating natural language to planning goals with large-language models,'' \emph{arXiv preprint arXiv:2302.05128}, 2023.

\bibitem{aeronautiques1998pddl}
C.~Aeronautiques, A.~Howe, C.~Knoblock, I.~D. McDermott, A.~Ram, M.~Veloso, D.~Weld, D.~W. Sri, A.~Barrett, D.~Christianson, \emph{et~al.}, ``{PDDL}| the planning domain definition language,'' \emph{Technical Report, Tech. Rep.}, 1998.

\bibitem{lifschitz2002answer}
V.~Lifschitz, ``Answer set programming and plan generation,'' \emph{Artificial Intelligence}, vol. 138, no. 1-2, pp. 39--54, 2002.

\bibitem{ding2023task}
Y.~Ding, X.~Zhang, C.~Paxton, and S.~Zhang, ``Task and motion planning with large language models for object rearrangement,'' in \emph{2023 IEEE/RSJ International Conference on Intelligent Robots and Systems (IROS)}.\hskip 1em plus 0.5em minus 0.4em\relax IEEE, 2023, pp. 2086--2092.

\bibitem{shirai2024vision}
K.~Shirai, C.~C. Beltran-Hernandez, M.~Hamaya, A.~Hashimoto, S.~Tanaka, K.~Kawaharazuka, K.~Tanaka, Y.~Ushiku, and S.~Mori, ``Vision-language interpreter for robot task planning,'' in \emph{2024 IEEE International Conference on Robotics and Automation (ICRA)}.\hskip 1em plus 0.5em minus 0.4em\relax IEEE, 2024, pp. 2051--2058.

\bibitem{Matthias2023}
M.~Kerzel, P.~Allgeuer, E.~Strahl, N.~Frick, J.-G. Habekost, M.~Eppe, and S.~Wermter, ``{NICOL}: A neuro-inspired collaborative semi-humanoid robot that bridges social interaction and reliable manipulation,'' \emph{IEEE Access}, vol.~11, pp. 123\,531--123\,542, 2023.

\bibitem{liu2024deepseek}
A.~Liu, B.~Feng, B.~Xue, B.~Wang, B.~Wu, C.~Lu, C.~Zhao, C.~Deng, C.~Zhang, C.~Ruan, \emph{et~al.}, ``Deepseek-v3 technical report,'' \emph{arXiv preprint arXiv:2412.19437}, 2024.

\bibitem{deepseekr1}
\BIBentryALTinterwordspacing
DeepSeek-AI, D.~Guo, D.~Yang, H.~Zhang, J.~Song, R.~Zhang, R.~Xu, Q.~Zhu, \emph{et~al.}, ``{DeepSeek-R1}: Incentivizing reasoning capability in {LLMs} via reinforcement learning,'' 2025. [Online]. Available: \url{https://arxiv.org/abs/2501.12948}
\BIBentrySTDinterwordspacing

\bibitem{gerevini2020introduction}
A.~E. Gerevini, ``An introduction to the planning domain definition language ({PDDL}): Book review,'' \emph{Artificial Intelligence}, vol. 280, p. 103221, 2020.

\bibitem{jiang2019task}
Y.-q. Jiang, S.-q. Zhang, P.~Khandelwal, and P.~Stone, ``Task planning in robotics: an empirical comparison of {PDDL}-and {ASP}-based systems,'' \emph{Frontiers of Information Technology \& Electronic Engineering}, vol.~20, pp. 363--373, 2019.

\bibitem{fikes1971strips}
R.~E. Fikes and N.~J. Nilsson, ``{STRIPS}: A new approach to the application of theorem proving to problem solving,'' \emph{Artificial Intelligence}, vol.~2, no. 3-4, pp. 189--208, 1971.

\bibitem{helmert2006fast}
M.~Helmert, ``The fast downward planning system,'' \emph{Journal of Artificial Intelligence Research}, vol.~26, pp. 191--246, 2006.

\bibitem{torreno2014fmap}
A.~Torreno, E.~Onaindia, and O.~Sapena, ``{FMAP}: Distributed cooperative multi-agent planning,'' \emph{Applied Intelligence}, vol.~41, pp. 606--626, 2014.

\bibitem{Ding2020}
Y.~Ding, X.~Zhang, X.~Zhan, and S.~Zhang, ``Task-motion planning for safe and efficient urban driving,'' in \emph{2020 IEEE/RSJ International Conference on Intelligent Robots and Systems (IROS)}, 2020, pp. 2119--2125.

\bibitem{jiang2019multi}
Y.~Jiang, H.~Yedidsion, S.~Zhang, G.~Sharon, and P.~Stone, ``Multi-robot planning with conflicts and synergies,'' \emph{Autonomous Robots}, vol.~43, no.~8, pp. 2011--2032, 2019.

\bibitem{kaelbling2013integrated}
L.~P. Kaelbling and T.~Lozano-P{\'e}rez, ``Integrated task and motion planning in belief space,'' \emph{The International Journal of Robotics Research}, vol.~32, no. 9-10, pp. 1194--1227, 2013.

\bibitem{jiao2021efficient}
Z.~Jiao, Z.~Zhang, W.~Wang, D.~Han, S.-C. Zhu, Y.~Zhu, and H.~Liu, ``Efficient task planning for mobile manipulation: a virtual kinematic chain perspective,'' in \emph{2021 IEEE/RSJ International Conference on Intelligent Robots and Systems (IROS)}.\hskip 1em plus 0.5em minus 0.4em\relax IEEE, 2021, pp. 8288--8294.

\bibitem{huang2022language}
W.~Huang, P.~Abbeel, D.~Pathak, and I.~Mordatch, ``Language models as zero-shot planners: Extracting actionable knowledge for embodied agents,'' in \emph{International Conference on Machine Learning (ICML)}.\hskip 1em plus 0.5em minus 0.4em\relax PMLR, 2022, pp. 9118--9147.

\bibitem{Zhao24EnhancingZeroshot}
X.~Zhao, M.~Li, W.~Lu, C.~Weber, J.~H. Lee, K.~Chu, and S.~Wermter, ``Enhancing zero-shot chain-of-thought reasoning in large language models through logic,'' in \emph{Proceedings of the 2024 Joint International Conference on Computational Linguistics, Language Resources and Evaluation ({{LREC-COLING}} 2024)}.\hskip 1em plus 0.5em minus 0.4em\relax {ELRA and ICCL}, May 2024, pp. 6144--6166.

\bibitem{stechly2024self}
K.~Stechly, K.~Valmeekam, and S.~Kambhampati, ``On the self-verification limitations of large language models on reasoning and planning tasks,'' \emph{arXiv preprint arXiv:2402.08115}, 2024.

\bibitem{Wang2023}
Z.~Wang, S.~Cai, G.~Chen, A.~Liu, X.~Ma, Y.~Liang, and T.~CraftJarvis, ``Describe, explain, plan and select: interactive planning with large language models enables open-world multi-task agents,'' in \emph{Proceedings of the 37th International Conference on Neural Information Processing Systems (NeurIPS)}, ser. NIPS '23.\hskip 1em plus 0.5em minus 0.4em\relax Red Hook, NY, USA: Curran Associates Inc., 2023.

\bibitem{Rana2024}
K.~Rana, J.~Haviland, S.~Garg, J.~Abou-Chakra, I.~Reid, and N.~Suenderhauf, ``{SayPlan}: Grounding large language models using {3D} scene graphs for scalable robot task planning,'' in \emph{7th Annual Conference on Robot Learning (CoRL)}, 2024.

\bibitem{chen2024autotamp}
Y.~Chen, J.~Arkin, C.~Dawson, Y.~Zhang, N.~Roy, and C.~Fan, ``{AutoTAMP}: Autoregressive task and motion planning with {LLMs} as translators and checkers,'' in \emph{2024 IEEE International Conference on Robotics and Automation (ICRA)}.\hskip 1em plus 0.5em minus 0.4em\relax IEEE, 2024, pp. 6695--6702.

\bibitem{chen2022towards}
Y.~Chen, T.~Wu, S.~Wang, X.~Feng, J.~Jiang, Z.~Lu, S.~McAleer, H.~Dong, S.-C. Zhu, and Y.~Yang, ``Towards human-level bimanual dexterous manipulation with reinforcement learning,'' \emph{Advances in Neural Information Processing Systems (NeurIPS)}, vol.~35, pp. 5150--5163, 2022.

\bibitem{avigal2022speedfolding}
Y.~Avigal, L.~Berscheid, T.~Asfour, T.~Kr{\"o}ger, and K.~Goldberg, ``Speedfolding: Learning efficient bimanual folding of garments,'' in \emph{2022 IEEE/RSJ International Conference on Intelligent Robots and Systems (IROS)}.\hskip 1em plus 0.5em minus 0.4em\relax IEEE, 2022, pp. 1--8.

\bibitem{Smith2012}
C.~Smith, Y.~Karayiannidis, L.~Nalpantidis, X.~Gratal, P.~Qi, D.~V. Dimarogonas, and D.~Kragic, ``Dual arm manipulation—a survey,'' \emph{Robotics and Autonomous systems}, vol.~60, no.~10, pp. 1340--1353, 2012.

\bibitem{cox2009efficient}
J.~Cox and E.~Durfee, ``Efficient and distributable methods for solving the multiagent plan coordination problem,'' \emph{Multiagent and Grid Systems}, vol.~5, no.~4, pp. 373--408, 2009.

\bibitem{ghallab2004automated}
M.~Ghallab, D.~Nau, and P.~Traverso, \emph{Automated Planning: theory and practice}.\hskip 1em plus 0.5em minus 0.4em\relax Elsevier, 2004.

\bibitem{Hong24MetaGPTMeta}
S.~Hong, M.~Zhuge, J.~Chen, X.~Zheng, Y.~Cheng, J.~Wang, C.~Zhang, Z.~Wang, S.~K.~S. Yau, Z.~Lin, L.~Zhou, C.~Ran, L.~Xiao, C.~Wu, and J.~Schmidhuber, ``{{MetaGPT}}: Meta programming for a multi-agent collaborative framework,'' in \emph{The Twelfth International Conference on Learning Representations (ICLR)}, 2024.

\bibitem{Kannan24SMARTLLMSmart}
S.~S. Kannan, V.~L.~N. Venkatesh, and B.-C. Min, ``{{SMART-LLM}}: {{Smart Multi-Agent Robot Task Planning}} using {{Large Language Models}},'' in \emph{2024 {{IEEE}}/{{RSJ International Conference}} on {{Intelligent Robots}} and {{Systems}} ({{IROS}})}, Oct. 2024, pp. 12\,140--12\,147.

\bibitem{minderer2023scaling}
M.~Minderer, A.~Gritsenko, and N.~Houlsby, ``Scaling open-vocabulary object detection,'' \emph{Advances in Neural Information Processing Systems (NeurIPS)}, vol.~36, pp. 72\,983--73\,007, 2023.

\bibitem{kovacs2012multi}
D.~L. Kovacs \emph{et~al.}, ``A multi-agent extension of {PDDL3.1},'' in \emph{ICAPS 2012 Proceedings of the 3rd Workshop on the International Planning Competition (WS-IPC 2012)}, 2012, pp. 19--37.

\bibitem{unified_planning_2025}
\BIBentryALTinterwordspacing
A.~Micheli, A.~Bit-Monnot, G.~R{\"o}ger, E.~Scala, A.~Valentini, L.~Framba, A.~Rovetta, A.~Trapasso, L.~Bonassi, A.~E. Gerevini, L.~Iocchi, F.~Ingrand, U.~Köckemann, F.~Patrizi, A.~Saetti, I.~Serina, and S.~Stock, ``Unified planning: Modeling, manipulating and solving {AI} planning problems in python,'' \emph{SoftwareX}, vol.~29, p. 102012, 2025. [Online]. Available: \url{https://www.sciencedirect.com/science/article/pii/S2352711024003820}
\BIBentrySTDinterwordspacing

\bibitem{frances2018best}
G.~Frances, H.~Geffner, N.~Lipovetzky, and M.~Ramir{\'e}z, ``Best-first width search in the {IPC} 2018: Complete, simulated, and polynomial variants,'' \emph{IPC-9 Planner Abstracts}, pp. 23--27, 2018.

\bibitem{chen2024not}
X.~Chen, J.~Xu, T.~Liang, Z.~He, J.~Pang, D.~Yu, L.~Song, Q.~Liu, M.~Zhou, Z.~Zhang, \emph{et~al.}, ``Do {NOT} think that much for 2+ 3=? {On} the overthinking of o1-like {LLMs},'' \emph{arXiv preprint arXiv:2412.21187}, 2024.

\end{thebibliography}

\end{document}